\documentclass[conference]{IEEEtran}
\IEEEoverridecommandlockouts
\usepackage{cite}
\usepackage{amsmath,amssymb,amsfonts}
\usepackage{algorithmic}
\usepackage{graphicx}
\usepackage{textcomp}
\usepackage{xcolor}
\usepackage{booktabs}
\usepackage{array}
\usepackage{comment}
\def\BibTeX{{\rm B\kern-.05em{\sc i\kern-.025em b}\kern-.08em
    T\kern-.1667em\lower.7ex\hbox{E}\kern-.125emX}}
\usepackage{adjustbox}
\usepackage{hyperref}
\usepackage[capitalize]{cleveref}
\usepackage{authblk}
\begin{document}

\title{EEMS: Edge-Prompt Enhanced Medical Image Segmentation Based on Learnable Gating Mechanism}

\author[1]{Han Xia}
\author[1]{Quanjun Li}
\author[1]{Qian Li}
\author[2*]{Zimeng Li\thanks{* Corresponding authors: li\_zimeng@szpu.edu.cn, xuhangc@hzu.edu.cn}}
\author[1]{Hongbin Ye}
\author[3,4]{Yupeng Liu}
\author[5]{Haolun Li}
\author[6*]{Xuhang Chen\thanks{This work was supported in part by Shenzhen Medical Research Fund (Grant No. A2503006), in part by the National Natural Science Foundation of China (Grant No. 62501412 and 82300277), in part by Shenzhen Polytechnic University Research Fund (Grant No. 6025310023K), in part by Medical Scientific Research Foundation of Guangdong Province (Grant No. B2025610 and B2023012), in part by the National Key Laboratory of Space Intelligent Control (Grant No. HTKJ2025KL502005) and in part by Guangdong Basic and Applied Basic Research Foundation (Grant No. 2024A1515140010).}}

\affil[1]{School of Advanced Manufacturing, Guangdong University of Technology}
\affil[2]{School of Electronic and Communication Engineering, Shenzhen Polytechnic University, Shenzhen, China}
\affil[3]{Department of Cardiology, Guangdong Provincial People's Hospital (Guangdong Academy of Medical Sciences),\protect\\Southern Medical University, Guangzhou, China}
\affil[4]{Guangdong Cardiovascular Institute, Guangdong Provincial People's Hospital,\protect\\Guangdong Academy of Medical Sciences, Guangzhou, China}
\affil[5]{College of Automation, Nanjing University of Posts and Telecommunications, Nanjing, China}
\affil[6]{School of Computer Science and Engineering, Huizhou University, Huizhou, China}

\maketitle

\begin{abstract}
Medical image segmentation is vital for diagnosis, treatment planning, and disease monitoring but is challenged by complex factors like ambiguous edges and background noise. We introduce EEMS, a new model for segmentation, combining an Edge-Aware Enhancement Unit (EAEU) and a Multi-scale Prompt Generation Unit (MSPGU). EAEU enhances edge perception via multi-frequency feature extraction, accurately defining boundaries. MSPGU integrates high-level semantic and low-level spatial features using a prompt-guided approach, ensuring precise target localization. The Dual-Source Adaptive Gated Fusion Unit (DAGFU) merges edge features from EAEU with semantic features from MSPGU, enhancing segmentation accuracy and robustness. Tests on datasets like ISIC2018 confirm EEMS's superior performance and reliability as a clinical tool.

\end{abstract}

\begin{IEEEkeywords}
Medical image segmentation; Edge enhancement; Prompt-guided
\end{IEEEkeywords}

\section{Introduction}
\begin{figure}[htbp]
    \centering
    \includegraphics[width=1\linewidth]{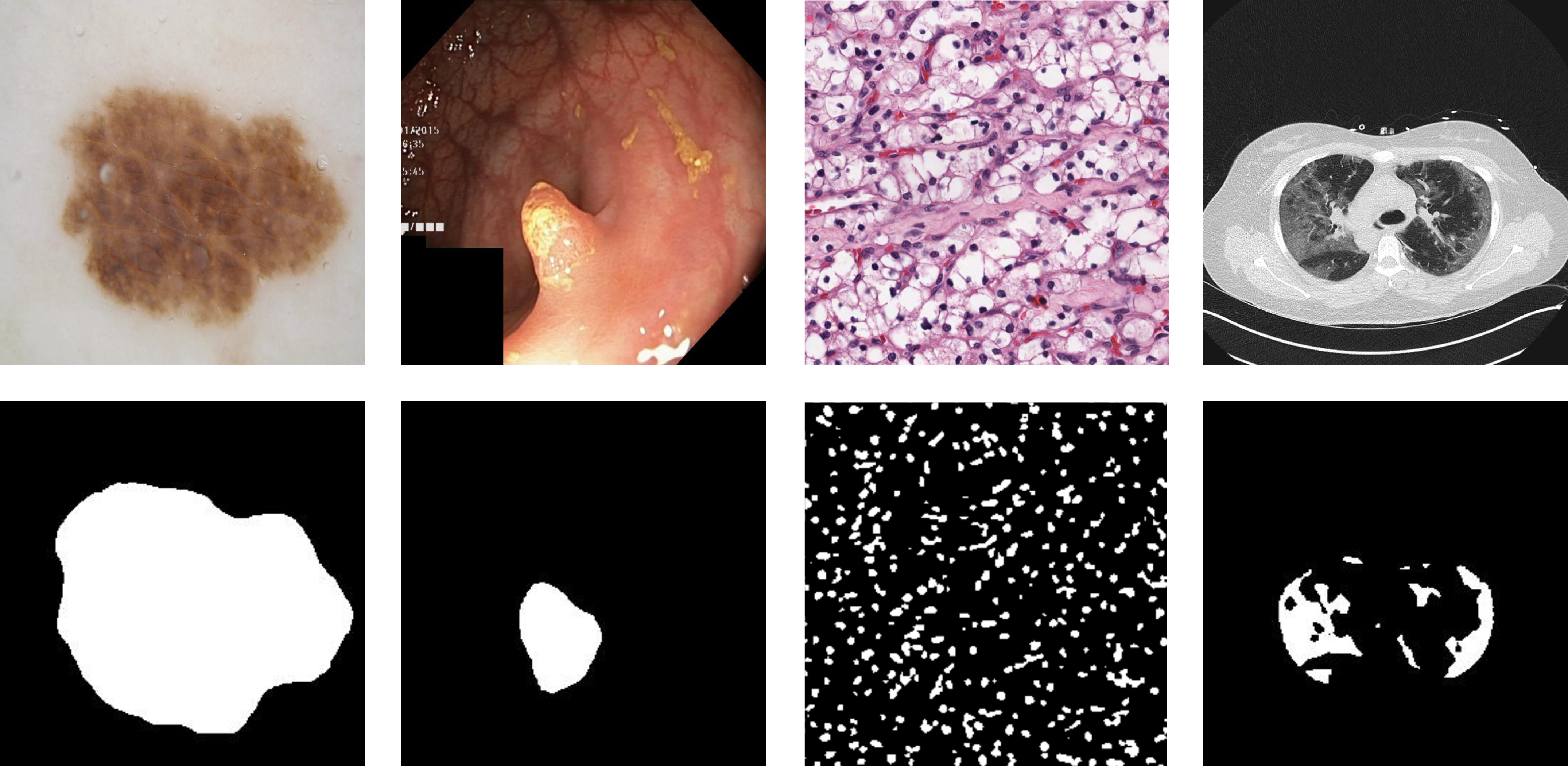}
    \caption{Effect of segmentation models applied to different body parts}
    \label{result}
\end{figure}

Medical image segmentation is vital in computer-aided diagnosis and treatment, designed to accurately distinguish anatomical structures or pathological areas from medical images. It is crucial for early disease diagnosis, treatment planning, and assessing treatment effectiveness. However, the challenges posed by low tissue contrast, blurred lesion edges, irregular shapes, and imaging noise make achieving precise, reliable automatic segmentation difficult. \Cref{result} illustrates the effect of medical image segmentation.

Recent advancements in deep learning \cite{li2025adaptive,li2022monocular,liu2024dh,liu2023coordfill,li2023cee,liu2024depth}, especially CNNs, have greatly improved medical image segmentation. U-Net \cite{ronneberger2015u} and its variants, known for their encoder-decoder structure and skip connections, effectively integrate multi-scale features and have become fundamental. However, challenges remain with complex images, particularly in fine edge segmentation and varied lesion shapes. Key issues include capturing multi-frequency information for better edge detail and using high-level semantic information for segmentation guidance.

To address these challenges, this paper proposes EEMS, a novel medical image segmentation model. EEMS's core innovation lies in its dual-branch structure, dedicated to edge enhancement and prompt guidance, with outputs adaptively fused by a Dual-Source Adaptive Gated Fusion Unit (DAGFU). Specifically, EEMS introduces:
\begin{itemize}
\item \textbf{Edge-Aware Enhancement Unit (EAEU):} This unit significantly enhances edge perception and precise boundary delineation through multi-frequency feature extraction and deformable feature refinement. It integrates multi-scale feature aggregation (with channel attention \cite{hu2018squeeze, woo2018cbam}) and deformable convolution \cite{dai2017deformable} to adaptively process edge information at various frequencies \cite{nam2024modality,liu2023explicit}.
\item \textbf{Multi-scale Prompt Generation Unit (MSPGU):} The MSPGU generates and integrates prompt information \cite{lei2024prompt} to guide the model's focus on key regions. It combines local and global context, innovatively employing a multi-scale prompt generation mechanism to produce discriminative prompts for accurate localization and segmentation in complex backgrounds.
\item \textbf{Dual-Source Adaptive Gated Fusion Unit (DAGFU):} DAGFU adaptively fuses features from EAEU and MSPGU using a learnable gating mechanism \cite{arevalo2017gated}, rather than simple concatenation. This dynamic strategy intelligently balances edge details and semantic prompts based on image content and task requirements.
\end{itemize}
Through the synergistic action of EAEU, MSPGU, and DAGFU, the EEMS model comprehensively utilizes image information, significantly improving segmentation accuracy and enhancing robustness to image complexity and noise.

\section{Methodology}

\subsection{EEMS Architecture Overview} 
\begin{figure}[htbp] 
    \centering
    \includegraphics[width=1\linewidth]{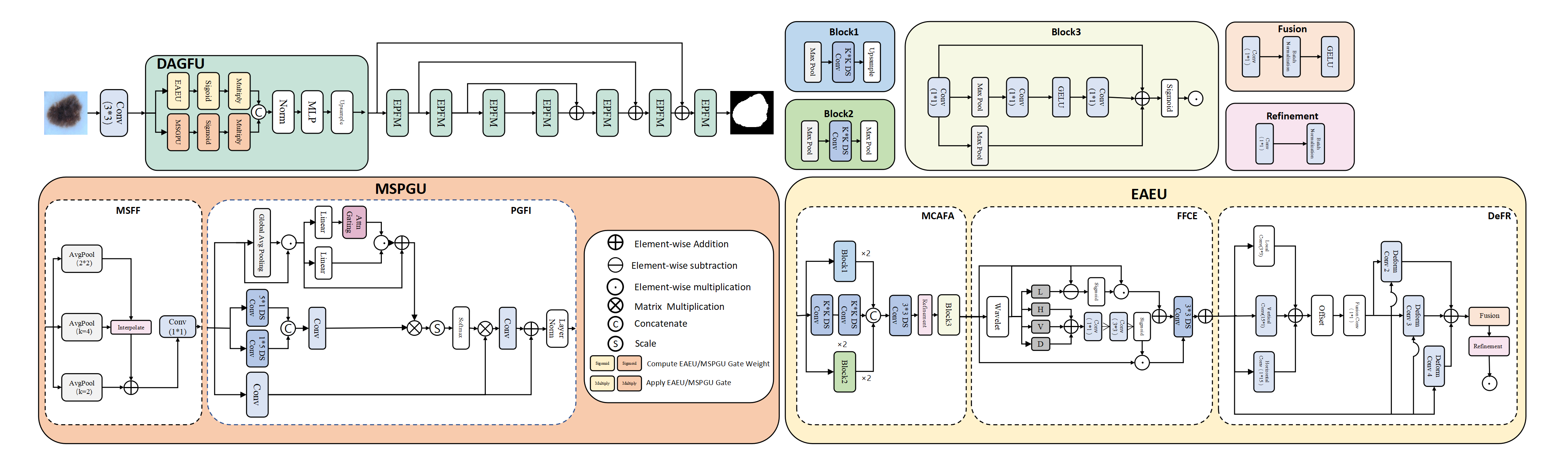} 
    \caption{EEMS Model Architecture Diagram}
    \label{fig:ehp_seg_architecture}
\end{figure}

EEMS adopts a classic encoder-decoder architecture, designed to effectively capture multi-scale and multi-frequency information in medical images. As shown in \cref{fig:ehp_seg_architecture}, the encoder progressively extracts high-level semantic features through a series of downsampling operations, while the EAEU module is responsible for enhancing edge information. The decoder progressively restores spatial resolution through upsampling operations and integrates low-level details by combining skip connections from the encoder. The MSPGU module generates prompt information at each stage, and the Dual-Source Adaptive Gated Fusion Unit (DAGFU) fuses the outputs of the EAEU and MSPGU, ultimately generating a refined segmentation mask.

\subsection{DAGFU: Dual-Source Adaptive Gated Fusion Unit} 
The Dual-Source Adaptive Gated Fusion Unit (DAGFU) is the core fusion component of EEMS, responsible for effectively fusing the edge-enhanced features generated by EAEU and the prompt-guided features generated by MSPGU. DAGFU is designed with MLP layers and up/downsampling operations to ensure seamless integration of features at different levels. Unlike traditional simple concatenation, DAGFU innovatively introduces a learnable gating mechanism, allowing the model to adaptively control the information flow from EAEU and MSPGU. By learning dynamic gating weights for each branch's features, DAGFU can intelligently decide which information should be adopted more, thereby achieving finer and more effective feature fusion.

Let $F_{EAEU}$ and $F_{MSPGU}$ be the output features of EAEU and MSPGU, respectively. The gating weights $G_{EAEU}$ and $G_{MSPGU}$ are calculated through $1 \times 1$ convolution and Sigmoid activation function:
\begin{equation}
G_{EAEU} = \sigma(\text{Conv}_{1 \times 1}(F_{EAEU})),
\end{equation}
\begin{equation}
G_{MSPGU} = \sigma(\text{Conv}_{1 \times 1}(F_{MSPGU})).
\end{equation}

Then, these gating weights are applied to the original features to obtain the gated features $F'_{EAEU}$ and $F'_{MSPGU}$:
\begin{equation}
F'_{EAEU} = F_{EAEU} \odot G_{EAEU},
\end{equation}
\begin{equation}
F'_{MSPGU} = F_{MSPGU} \odot G_{MSPGU}.
\end{equation}

Finally, the final output of DAGFU, $F_{DAGFU\_out}$, is obtained by concatenating the gated features and passing them through a Multi-Layer Perceptron (MLP) layer for fusion:
\begin{equation}
F_{DAGFU\_out} = \text{MLP}(\text{Concat}(F'_{EAEU}, F'_{MSPGU})),
\end{equation}
where $\text{Concat}$ denotes concatenation along the channel dimension, and $\text{MLP}$ represents the sequence of $1 \times 1$ convolutions and normalization layers used for feature transformation within DAGFU.

\subsection{EAEU: Edge-Aware Enhancement Unit} 
The Edge-Aware Enhancement Unit (EAEU) aims to significantly enhance the model's perception of image edge details through multi-frequency feature extraction and refinement. The core idea of EAEU is to utilize multi-scale feature aggregation and deformable convolution to adapt to irregular edges. EAEU mainly consists of the following three sub-modules:

\subsubsection{FFCE: Frequency Feature Combination and Enhancement} 

The Frequency Feature Combination and Enhancement (FFCE) module utilizes wavelet transform to decompose the image into different frequency components, where high-frequency components are rich in edge and texture information. By processing and fusing these multi-frequency components, FFCE can enhance edge features from the frequency domain, providing the model with more comprehensive detailed information.
Let $X$ be the input feature of FFCE. Through wavelet transform, we can obtain low-frequency component $L$ and high-frequency components $LH, HL, HH$.
First, the high-frequency components $LH, HL, HH$ are fused through concatenation and a $1 \times 1$ convolution $\Phi_{HF}$, yielding the fused high-frequency feature $H_{fused}$:
\begin{equation}
\begin{aligned}
    H_{fused} &= \Phi_{HF}(\text{Concat}(\text{Upsample}(LH),\\& \text{Upsample}(HL), \text{Upsample}(HH))),
\end{aligned}
\end{equation}
where $\text{Upsample}$ denotes the upsampling operation.
Next, high-frequency attention feature $x_1$ and low-frequency difference attention feature $x_2$ are calculated through the Sigmoid activation function and element-wise multiplication:
\begin{equation}
x_1 = \sigma(\text{Conv}_{1 \times 1}(H_{fused})) \odot X,
\end{equation}
\begin{equation}
x_2 = \sigma(X - L) \odot X.
\end{equation}

Then, $x_1$ and $x_2$ are concatenated and passed through a fusion module $\Phi_{Final}$ to obtain the final fused feature $F_{final}$:
\begin{equation}
F_{final} = \Phi_{Final}(\text{Concat}(x_1, x_2)),
\end{equation}
where $\Phi_{Final}$ is a fusion module consisting of $1 \times 1$ convolution, batch normalization, and GELU activation.
Finally, the output of FFCE, $F_{FFCE\_out}$, is obtained by passing $F_{final}$ through a refinement module $\text{Refinement}$ and then performing a residual connection with the original input $X$:
\begin{equation}
F_{FFCE\_out} = X + \text{Refinement}(F_{final}).
\end{equation}

\subsubsection{MCAFA: Multi-Scale Channel Attention Feature Aggregation} 

The Multi-Scale Channel Attention Feature Aggregation (MCAFA) module utilizes convolutional kernels of different sizes and sampling operations (such as upsampling and max pooling) to capture image features at multiple scales. It enhances the perception of multi-scale edges by processing information from different receptive fields in parallel. To further optimize the aggregation effect of multi-scale features, we innovatively introduced a channel attention mechanism in the MCAFA module. This mechanism can adaptively learn and emphasize the importance of different-scale feature channels, effectively suppressing redundant information, thereby generating more discriminative and informative fused features, greatly enhancing the model's ability to capture fine edge details.
Its mathematical representation is as follows:
\begin{equation}
\begin{aligned}
M_c(F) &= \sigma(\text{Conv}_{1 \times 1}(\text{ReLU}(\text{Conv}_{1 \times 1}(\text{AvgPool}(F))))\\&+ \text{Conv}_{1 \times 1}(\text{ReLU}(\text{Conv}_{1 \times 1}(\text{MaxPool}(F))))),
\end{aligned}
\end{equation}
where $F$ is the input feature map, $\text{AvgPool}$ and $\text{MaxPool}$ represent global average pooling and global max pooling respectively, $\text{Conv}_{1 \times 1}$ is a $1 \times 1$ convolution, $\text{ReLU}$ is the activation function, and $\sigma$ is the Sigmoid activation function. Finally, the channel attention-weighted feature $F_{MCAFA\_out}$ is:
\begin{equation}
F_{MCAFA\_out} = M_c(F_{extracted}) \odot F_{extracted},
\end{equation}
where $F_{extracted}$ is the feature after initial multi-scale feature concatenation and extraction layers in the MCAFA module, and $\odot$ denotes element-wise multiplication.

\subsubsection{DeFR: Deformable Feature Refinement} 
The Deformable Feature Refinement (DeFR) module is a key component of EAEU, which utilizes deformable convolution to adapt to the geometric deformation of target objects. Unlike traditional convolutions, deformable convolutions learn offsets for each sampling point, allowing the convolutional kernel to adaptively adjust its receptive field, thereby more precisely capturing irregular edges and complex lesion structures. This is particularly important for lesions with varied morphologies in medical images, ensuring the accuracy of edge segmentation.
Let $F_{in}$ be the input feature of DeFR. First, the offset $\Delta_o$ is generated by the directional offset generator $\Psi_{offset}$:
\begin{equation}
\Delta_o = \Psi_{offset}(\text{Concat}(O_{3,3}, O_{15,1}, O_{1,15})),
\end{equation}
where $O_{k_x, k_y}$ are offset features obtained through convolutions with different receptive fields.
Then, the input feature $F_{in}$ and offset $\Delta_o$ are fed into deformable convolutions $DCN^d$ with different dilation rates. The outputs $D_d$ of all deformable convolutions are concatenated and passed through a fusion layer $\Phi_{fusion}$ to obtain the fused deformable feature $F_{deform}$:
\begin{equation}
F_{deform} = \Phi_{fusion}(\text{Concat}(D_2, D_3, D_4)),
\end{equation}
where $\Phi_{fusion}$ is a fusion module consisting of $1 \times 1$ convolution, batch normalization, and GELU activation.
Finally, the output of DeFR, $F_{DeFR\_out}$, is obtained by passing $F_{deform}$ through a refinement module $\Omega$ and then performing a residual connection with the original input $F_{in}$:
\begin{equation}
F_{DeFR\_out} = F_{in} + \Omega(F_{deform}).
\end{equation}

\subsection{MSPGU: Multi-scale Prompt Generation Unit} 
The Multi-scale Prompt Generation Unit (MSPGU) aims to generate and integrate prompt information, enabling the model to more effectively integrate high-level semantic information with low-level spatial features, thereby achieving accurate localization and segmentation of target regions in complex backgrounds. MSPGU combines local and global prompt generation mechanisms, mainly including the following four sub-modules:

\subsubsection{MSFF: Multi-Scale Feature Fusion} 
MSPGU innovatively introduces the MSFF method, which fuses feature information from different resolutions when generating low-frequency and high-frequency prompts. Specifically, let $X$ be the input feature of MSPGU. Through average pooling operations at different scales, we obtain multi-scale features $P_0, P_k$:
\begin{equation}
P_0 = \text{AvgPool}_{\text{kernel}=2}(X),
\end{equation}
\begin{equation}
P_k = \text{AvgPool}_{\text{kernel}=s_k}(X), \quad \forall s_k \in \{2, 4, \dots\},
\end{equation}
where $s_k$ is the pooling kernel size defined in \texttt{multi\_scale\_pool}. To fuse these features, we upsample them to a uniform size (e.g., the size of $P_0$):
\begin{equation}
P'_k = \text{Interpolate}(P_k, \text{size}(P_0)).
\end{equation}

Then, the multi-scale fused feature $F_{MSFF\_fused}$ is obtained by concatenation and $1 \times 1$ convolution:
\begin{equation}
F_{MSFF\_fused} = \text{Conv}_{1 \times 1}(\text{Concat}(P_0, P'_1, P'_2, \dots)),
\end{equation}
where $F_{MSFF\_fused}$ will serve as input for Low-Frequency Prompt Generation ($LPG$) and High-Frequency Prompt Generation ($HPG$).

\subsubsection{PGFI: Prompt-Guided Feature Integration} 
The PGFI module is the core of MSPGU, which fuses the generated low-frequency prompt $P_L$, high-frequency prompt $P_H$, and original pooled feature $P_0$ to generate the final prompt-guided feature $F_{MSPGU\_out}$ through a cross-attention mechanism.
\begin{equation}
Q = P_H,
\end{equation}
\begin{equation}
K = P_L,
\end{equation}
\begin{equation}
V = \text{Linear}_O(P_0),
\end{equation}
where the attention weight $A = \text{Softmax}(Q K^T / \sqrt{d_k})$, where $d_k$ is the feature dimension.
The final prompt-guided feature $F_{MSPGU\_out}$ is calculated as:
\begin{equation}F_{MSPGU\_out} = \text{Upsample}(\text{LayerNorm}(\text{Linear}_P(A V) + V))\end{equation}
Where $\text{Linear}_O$ and $\text{Linear}_P$ are linear projection layers, $\text{LayerNorm}$ is layer normalization, and $\text{Upsample}$ is the upsampling operation.

\section{Experiments}

\subsection{Datasets and Evaluation Metrics}
To comprehensively evaluate the EEMS model's effectiveness, we conducted extensive experiments on four public datasets: ISIC2018 \cite{ISIC2018}, Kvasir \cite{Kvasir}, Monu-Seg \cite{Monu-Seg}, and COVID-19 \cite{COVID-19}. Comparisons were made against several existing state-of-the-art models.



\subsection{Implementation Details}
All models were implemented in PyTorch and trained on a computing platform equipped with NVIDIA 4090 GPUs.
\textbf{Optimizer:} We utilized the AdamW optimizer with an initial learning rate of 0.001.
\textbf{Learning Rate Scheduler:} A CosineAnnealingLR scheduler was employed, with $T_{max}=50$ and $eta_{min}=0.00001$.
\textbf{Loss Function:} A BCE-Dice hybrid loss function was used. Its weights were dynamically adjusted during training ($lambda_{start}=0.7$, $lambda_{end}=0.3$, $lambda_{decay-epochs}=100$) to balance pixel-level classification and region overlap optimization.
\textbf{Training Epochs:} Models were trained for 200 epochs.
\textbf{Batch Size:} A batch size of 8 was used.
\textbf{Image Size:} A progressive training strategy was adopted: 100 epochs with $128\times128$ image size, followed by 100 epochs with $256\times256$ image size.
\textbf{Data Augmentation:} Standard data augmentation techniques, including random rotation, random vertical flip, random horizontal flip, and normalization, were applied.

\subsection{Results and Analysis}
To comprehensively evaluate the effectiveness of the EEMS model, we assessed its performance on the ISIC2018, Kvasir, Monu-Seg, and COVID-19 datasets, and compared it with several existing state-of-the-art models.

\subsubsection{Performance on ISIC2018 Dataset}
\cref{tab:isic2018_comparison} presents the detailed quantitative comparison results of the EEMS model with existing SOTA methods on the ISIC2018 dataset.

\begin{table}[ht]
\centering
\footnotesize
\caption{Performance Comparison of EEMS with Existing SOTA Methods on ISIC2018 Dataset}
\label{tab:isic2018_comparison}
\setlength{\tabcolsep}{3pt}
\renewcommand{\arraystretch}{1.0}
\adjustbox{width=\linewidth}{
\begin{tabular}{@{} *{10}{>{\centering\arraybackslash}p{1.5cm}} @{}}
\toprule
\textbf{Metric} & \textbf{U-Net}\cite{ronneberger2015u} & \textbf{MALU-Net}\cite{MALU-Net} & \textbf{UltraLight-VMUNet}\cite{UltraLightVM-UNet} & \textbf{EGE-UNet}\cite{EGE-UNet} & \textbf{VPTTA}\cite{VPTTA} & \textbf{SAM-Med2D}\cite{SAM-Med2D} & \textbf{MLWNet}\cite{MLWNet} & \textbf{EMCAD}\cite{EMCAD} & \textbf{EEMS (Ours)} \\
\midrule
mIoU        & 0.8004 & 0.7976 & 0.8110 & 0.8108 & 0.7842 & 0.7383 & 0.7650 & 0.8071 & \textbf{0.8498} \\
Dice        & 0.8955 & 0.8971 & 0.8988 & 0.8909 & 0.8956 & 0.8613 & 0.8494 & 0.8997 & \textbf{0.9188} \\
Accuracy    & 0.9404 & 0.9513 & 0.9549 & 0.9495 & 0.9527 & 0.9514 & 0.9535 & 0.9531 & \textbf{0.9652} \\
Specificity & 0.9768 & 0.9738 & 0.9699 & 0.9667 & 0.9662 & 0.9742 & 0.9721 & 0.9750 & \textbf{0.9787} \\
Sensitivity & 0.8161 & 0.8730 & 0.8907 & 0.8896 & 0.9061 & 0.8738 & 0.8891 & 0.8760 & \textbf{0.9160} \\
\bottomrule
\end{tabular}}
\end{table}

\begin{figure}[h]
    \centering
    \includegraphics[width=1\columnwidth]{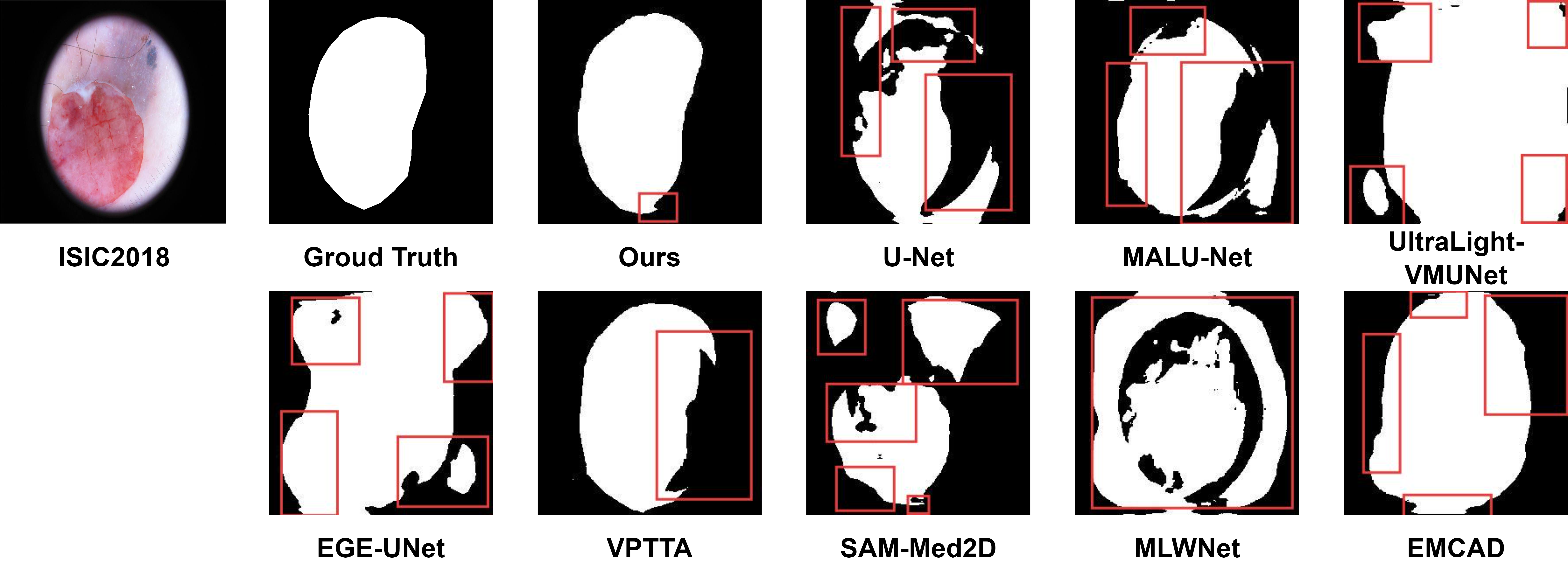}
    \caption{EEMS Model Visual Segmentation Results}
    \label{fig:visual_comparison}
\end{figure}

\cref{tab:isic2018_comparison} clearly demonstrates EEMS's superior performance on the ISIC2018 dataset, surpassing current SOTA models across all metrics. EEMS achieved a remarkable mIoU of \textbf{0.8498}, indicating its enhanced capability in accurately identifying target areas through superior overlap. Its Dice coefficient of \textbf{0.9188} further confirms robust pixel-level segmentation accuracy, showing a significant lead over EMCAD (0.8997) and generally outperforming other models. EEMS also exhibited a notable advantage in overall accuracy (\textbf{0.9652}) and specificity (\textbf{0.9787}), with high specificity highlighting its exceptional precision in identifying background pixels, crucial for minimizing false positives. Furthermore, a sensitivity of \textbf{0.9160} indicates effective identification of true lesion pixels, reducing false negatives and notably outperforming current SOTA models. Collectively, EEMS strikes an excellent balance between accuracy and sensitivity, reinforced by its strong specificity. \Cref{fig:visual_comparison} presents EEMS's segmentation predictions on representative ISIC2018 images.

From the visualization results, it can be observed that the EEMS model can generate relatively smooth and accurate segmentation masks, showing good adaptability, especially when dealing with images with blurred lesion boundaries or irregular shapes.
These results strongly demonstrate that the EEMS model, through gated feature fusion in the EPMF module, dynamically fuses the output results of EAEU and MSPGU, enabling more comprehensive utilization of image information, improving segmentation accuracy, and enhancing the model's robustness to image complexity and noise, thereby achieving excellent performance in medical image segmentation tasks.

\subsubsection{Performance on Kvasir-SEG Dataset}
\cref{tab:kvasir_comparison} presents the detailed quantitative comparison results of the EEMS model with existing SOTA methods on the Kvasir-SEG dataset.

\begin{table}[ht]
\centering
\footnotesize
\caption{Performance Comparison of EEMS with Existing SOTA Methods on Kvasir-SEG Dataset}
\label{tab:kvasir_comparison}
\setlength{\tabcolsep}{3pt}
\renewcommand{\arraystretch}{1.0}
\adjustbox{width=\linewidth}{
\begin{tabular}{@{} *{10}{>{\centering\arraybackslash}p{1.5cm}} @{}}
\toprule
\textbf{Metric} & \textbf{U-Net}\cite{ronneberger2015u} & \textbf{MALU-Net}\cite{MALU-Net} &\textbf{UltraLight-VMUNet}\cite{UltraLightVM-UNet} & \textbf{EGE-UNet}\cite{EGE-UNet} & \textbf{VPTTA}\cite{VPTTA} & \textbf{SAM-Med2D}\cite{SAM-Med2D} & \textbf{MLWNet}\cite{MLWNet} & \textbf{EMCAD}\cite{EMCAD} & \textbf{EEMS (Ours)} \\
\midrule
mIoU        & 0.7330 & 0.5351 & 0.6100 & 0.5604 & 0.5164 & 0.5911 & 0.6636 & 0.7173 & \textbf{0.8121} \\
DSC         & 0.8459 & 0.6972 & 0.7577 & 0.7182 & 0.6811 & 0.7430 & 0.7977 & 0.8353 & \textbf{0.8963} \\
Accuracy    & 0.9253 & 0.8998 & 0.9190 & 0.9108 & 0.9076 & \textbf{0.9964} & 0.9076 & 0.9478 & 0.9697 \\
Specificity & 0.9786 & 0.9348 & 0.9441 & 0.9499 & 0.9639 & \textbf{0.9964} & 0.9639 & 0.9716 & 0.9811 \\
Sensitivity & 0.8145 & 0.7170 & 0.7879 & 0.7068 & 0.6137 & 0.6020 & 0.7620 & 0.8234 & \textbf{0.9026} \\
\bottomrule
\end{tabular}}
\end{table}

On the Kvasir-SEG dataset, EEMS achieved the best performance in mIoU, Dice, and Sensitivity metrics, especially reaching \textbf{0.8121} in mIoU, significantly outperforming other comparative methods. This indicates that EEMS has excellent accuracy and robustness in colonoscopy image polyp segmentation tasks.

\subsubsection{Performance on Monu-Seg Dataset}
\cref{tab:monu_seg_comparison} presents the detailed quantitative comparison results of the EEMS model with existing SOTA methods on the Monu-Seg dataset.

\begin{table}[ht]
\centering
\footnotesize
\caption{Performance Comparison of EEMS with Existing SOTA Methods on Monu-Seg Dataset}
\label{tab:monu_seg_comparison}
\setlength{\tabcolsep}{3pt}
\renewcommand{\arraystretch}{1.0}
\adjustbox{width=\linewidth}{
\begin{tabular}{@{} *{10}{>{\centering\arraybackslash}p{1.5cm}} @{}}
\toprule
\textbf{Metric} & \textbf{U-Net}\cite{ronneberger2015u} & \textbf{MALU-Net}\cite{MALU-Net} &\textbf{UltraLight-VMUNet}\cite{UltraLightVM-UNet} & \textbf{EGE-UNet}\cite{EGE-UNet} & \textbf{VPTTA}\cite{VPTTA} & \textbf{SAM-Med2D}\cite{SAM-Med2D} & \textbf{MLWNet}\cite{MLWNet} & \textbf{EMCAD}\cite{EMCAD} & \textbf{EEMS (Ours)} \\
\midrule
mIoU        & \textbf{0.6784} & 0.5133 & 0.5600 & 0.5009 & 0.4151 & 0.2699 & 0.6535 & 0.5603 & 0.6543 \\
DSC         & \textbf{0.8084} & 0.6784 & 0.7180 & 0.6674 & 0.5867 & 0.4250 & 0.7904 & 0.7182 & 0.7910 \\
Accuracy    & 0.9394 & \textbf{0.9134} & 0.9094 & 0.8873 & 0.8662 & 0.9433 & 0.9332 & 0.9263 & 0.9129 \\
Specificity & 0.9640 & 0.9406 & 0.9406 & 0.9386 & 0.9188 & \textbf{0.9977} & 0.9540 & 0.9522 & 0.9367 \\
Sensitivity & \textbf{0.8587} & 0.7248 & 0.7489 & 0.7351 & 0.5911 & 0.2774 & 0.8187 & 0.7461 & 0.8187 \\
\bottomrule
\end{tabular}}
\end{table}

On the Monu-Seg dataset, EEMS performed excellently in mIoU, Dice, and Sensitivity metrics, especially reaching \textbf{0.8187} in Sensitivity, indicating its strong ability to identify cell nuclei. Although slightly lower than U-Net in mIoU and Dice, its various metrics are at the upper level among current SOTA models, demonstrating its outstanding performance in nucleus segmentation tasks.

\subsubsection{Performance on COVID-19 Dataset}
\cref{tab:covid19_comparison} presents the detailed quantitative comparison results of the EEMS model with existing SOTA methods on the COVID-19 dataset.

\begin{table}[ht]
\centering
\footnotesize
\caption{Performance Comparison of EEMS with Existing SOTA Methods on COVID-19 Dataset}
\label{tab:covid19_comparison}
\setlength{\tabcolsep}{3pt}
\renewcommand{\arraystretch}{1.0}
\adjustbox{width=\linewidth}{
\begin{tabular}{@{} *{10}{>{\centering\arraybackslash}p{1.5cm}} @{}}
\toprule
\textbf{Metric} & \textbf{U-Net}\cite{ronneberger2015u} & \textbf{MALU-Net}\cite{MALU-Net} &\textbf{UltraLight-VMUNet}\cite{UltraLightVM-UNet} & \textbf{EGE-UNet}\cite{EGE-UNet} & \textbf{VPTTA}\cite{VPTTA} & \textbf{SAM-Med2D}\cite{SAM-Med2D} & \textbf{MLWNet}\cite{MLWNet} & \textbf{EMCAD}\cite{EMCAD} & \textbf{EEMS (Ours)} \\
\midrule
mIoU        & 0.3605 & 0.4811 & 0.5532 & 0.3912 & 0.4591 & 0.4025 & 0.4295 & 0.4120 & \textbf{0.7073} \\
DSC         & 0.5300 & 0.6497 & 0.7123 & 0.5624 & 0.6293 & 0.5739 & 0.6009 & 0.5835 & \textbf{0.8286} \\
Accuracy    & 0.9784 & 0.9845 & 0.9867 & 0.9805 & 0.9859 & 0.9856 & 0.9808 & 0.9790 & \textbf{0.9915} \\
Specificity & 0.9881 & 0.9932 & 0.9933 & 0.9909 & 0.9968 & \textbf{0.9981} & 0.9872 & 0.9966 & 0.9941 \\
Sensitivity & 0.5488 & 0.6176 & 0.7092 & 0.5397 & 0.5203 & 0.4359 & 0.6221 & 0.6334 & \textbf{0.8808} \\
\bottomrule
\end{tabular}}
\end{table}

On the COVID-19 dataset, EEMS achieved the best performance in mIoU, DSC, Accuracy, and Sensitivity metrics. Its performance in mIoU, DSC, and Sensitivity reached \textbf{0.7073}, \textbf{0.8286}, and \textbf{0.8808} respectively, which are significantly ahead of current SOTA models. This indicates that EEMS has powerful capabilities in lung lesion segmentation tasks and can effectively process complex CT images.

\subsection{Ablation Studies}
\subsubsection{Analysis of Module Effectiveness}
To validate the effectiveness of each proposed module within the EEMS model and their respective contributions to overall performance, we conducted a series of ablation experiments. Ablation studies systematically evaluate the importance of each component by observing changes in model performance after removing specific components or sub-modules.

In this study, our primary focus is on the contributions of EAEU, MSPGU, and DAGFU, as well as their key internal sub-modules. Specifically, we assessed their impact by removing the following components:
\begin{itemize}
    \item \textbf{EAEU Component Ablation:} Separately removing the MCAFA, FFCE, and DeFR modules.
    \item \textbf{MSPGU Component Ablation:} Separately removing the MSFF and PGFI modules.
    \item \textbf{DAGFU Component Ablation:} Removing the learnable gating mechanism in DAGFU, which is responsible for adaptively fusing features from both the EAEU and MSPGU branches.
\end{itemize}
After each removal, we re-trained and re-evaluated the model on the ISIC2018 dataset, recording changes in various performance metrics. These experiments aim to quantify the specific contribution of each innovative aspect to the model's final segmentation accuracy and robustness.

\begin{figure}[htbp]
    \centering
    \includegraphics[width=1\columnwidth]{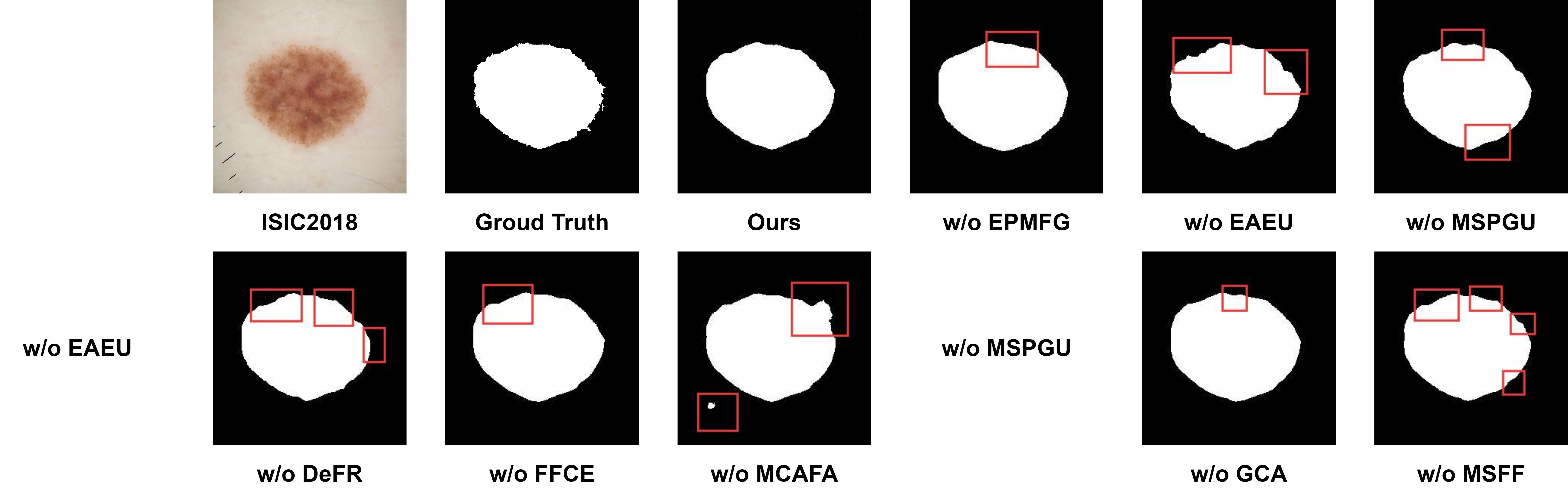}
    \caption{Ablation experiment results of EEMS on ISIC2018 dataset}
    \label{abligation}
\end{figure}

\Cref{abligation} visually presents the results of these ablation experiments, clearly illustrating the critical role each module plays in enhancing the performance of the EEMS model.

\subsubsection{Ablation experiment of learnable gating mechanism in the DAGFU}
To verify the effectiveness of the learnable gating mechanism, we conducted an ablation study where we removed it from the DAGFU module and instead used simple feature concatenation for fusion. \cref{tab:ablation_gated_fusion} shows the performance changes of the model on the ISIC2018 dataset after removing the learnable gating mechanism. 

\begin{table}[h!]
\centering
\caption{learnable gating mechanism in DAGFU Ablation Study}
\label{tab:ablation_gated_fusion}
\begin{tabular}{@{}lccc@{}}
\toprule
\textbf{Model Configuration} & \textbf{mIoU} & \textbf{Dice} & \textbf{Accuracy} \\
\midrule
EEMS(Origin)  & \textbf{0.8498} & \textbf{0.9188} & \textbf{0.9656} \\
EEMS (w/o learnable gating mechanism) & 0.8212 & 0.9018 & 0.9587 \\ 
\bottomrule
\end{tabular}
\end{table}

As shown in \cref{tab:ablation_gated_fusion}, when the learnable gating mechanism is removed, the model's performance decreased notably, with mIoU dropping from \textbf{0.8304} to \textbf{0.8250} and Dice from \textbf{0.9074} to \textbf{0.9020}. This strongly proves the important role of the learnable gating mechanism in DAGFU in adaptively controlling and optimizing the fusion process of edge-enhanced features and prompt-guided features. It can dynamically adjust the information flow according to feature discriminability and task requirements, thereby achieving finer and more effective feature integration, and further improving overall segmentation performance.

\subsubsection{Ablation experiment of EAEU and MSPGU}
To deeply analyze the contribution of EAEU and MSPGU, both individually and synergistically, to model performance, we conducted detailed ablation experiments. We compared model configurations using only EAEU, only MSPGU, and both EAEU and MSPGU simultaneously. \cref{EAEU and MSPGU Effectiveness Ablation} shows the performance of these configurations on the ISIC2018 dataset.

\begin{table}[h!]
\centering
\caption{EAEU and MSPGU Effectiveness Ablation Study based on ISIC2018 Dataset}
\label{EAEU and MSPGU Effectiveness Ablation}
\begin{tabular}{@{}lccc@{}}
\toprule
\textbf{Model Configuration} & \textbf{mIoU} & \textbf{Dice} & \textbf{Accuracy} \\
\midrule
EAEU Only           & 0.8176        & 0.8996       & 0.9569          \\
MSPGU Only           & 0.8262        & 0.9048        & 0.9595            \\
EEMS (EAEU + MSPGU) & \textbf{0.8498} & \textbf{0.9188} & \textbf{0.9656} \\
\bottomrule
\end{tabular}
\end{table}

From \cref{EAEU and MSPGU Effectiveness Ablation}, it can be seen that when both EAEU and MSPGU are used simultaneously, the model achieved the best performance across all metrics. This indicates that the synergistic effect of EAEU focusing on edge enhancement and MSPGU focusing on prompt guidance can significantly improve the model's overall segmentation capability.

\subsubsection{Ablation experiment of EAEU and MSPGU Sub-modules}
By conducting an ablation study on the effectiveness of each internal component of EAEU and MSPGU, we can quantify the importance of each sub-module. \cref{tab:ablation_components} shows the performance changes of the model on ISIC2018 after removing specific components.

\begin{table}[ht]
\centering
\caption{EAEU and MSPGU Internal Component Effectiveness Ablation Study (based on ISIC2018 Dataset)}
\label{tab:ablation_components}
\begin{tabular}{@{}lccc@{}}
\toprule
\textbf{Model Configuration} & \textbf{mIoU} & \textbf{Dice} & \textbf{Accuracy} \\
\midrule
EEMS(Origin)  & \textbf{0.8498} & \textbf{0.9188} & \textbf{0.9656}  \\
\midrule
\multicolumn{4}{l}{\textit{MSPGU Component Ablation}} \\
\quad w/o MSFF     & 0.8327        & 0.9087        & 0.9618            \\ 
\quad w/o PGFI     & 0.8331        & 0.9089        & 0.9608           \\ 
\midrule
\multicolumn{4}{l}{\textit{EAEU Component Ablation}} \\
\quad w/o MCAFA     & 0.8422        & 0.9143        & 0.9629            \\
\quad w/o FFCE     & 0.8316        & 0.9080        & 0.9594            \\
\quad w/o DeFR      & 0.8372        & 0.9113        & 0.9618            \\
\bottomrule
\end{tabular}
\end{table}

As shown in \cref{tab:ablation_components}, removing any single component leads to a decrease in model performance, indicating that each sub-module within EAEU and MSPGU is crucial. In particular, the most significant drop in mIoU occurred when PGFI was removed, emphasizing the core role of the PGFI mechanism in integrating multi-source information and improving segmentation accuracy.

\section{Conclusion}
This paper presents EEMS, a novel model for medical image segmentation, combining the Edge-Aware Enhancement Unit (EAEU) and Multi-scale Prompt Generation Unit (MSPGU). EAEU improves edge perception through multi-frequency feature refinement, while MSPGU enhances high-level semantic integration using a prompt-guided mechanism. Key innovations include channel attention in MCAFA, gated fusion in DAGFU, and multi-scale prompt generation in MSPGU, leading to notable performance gains. Validated on ISIC2018, Kvasir, Monu-Seg, and COVID-19 datasets, EEMS excels in complex edge and target region segmentation. Future work will optimize EAEU, MSPGU, explore advanced fusion strategies, and test EEMS's generalization on varied datasets.

\bibliographystyle{IEEEtran}
\bibliography{ref}

\end{document}